\documentclass[journal,transmag]{IEEEtran}
\usepackage{graphicx}
\usepackage{multirow}
\usepackage{subfig}
\usepackage{float}
\usepackage{enumerate}
\usepackage{amsfonts}
\usepackage{amsmath}
\usepackage{algorithm}
\usepackage{algorithmicx}
\usepackage{algpseudocode}
\usepackage{textcomp}
\ifCLASSINFOpdf
\else
\fi
\begin{document}
\title{Unsupervised Feature Learning by Autoencoder and Prototypical Contrastive Learning for Hyperspectral Classification}


\author{\IEEEauthorblockN{Zeyu Cao\IEEEauthorrefmark{1},
Xiaorun Li\IEEEauthorrefmark{1},
Liaoying Zhao\IEEEauthorrefmark{2}}
%\IEEEauthorblockA{\IEEEauthorrefmark{1}Zhejiang University, College of Electrical Engineering, 866 Yuhangtang Rd, Hangzhou, China, 310058}
%\IEEEauthorblockA{\IEEEauthorrefmark{2}HangZhou Dianzi University, China Institute of Computer Application Technology, Xiasha Higher Education Zone, Hangzhou, China, 310018}

\thanks{Manuscript received December 1, 2012; revised August 26, 2015. 
Corresponding author: Xiaorun Li (email:lxr@zju.edu.cn).
\IEEEauthorrefmark{1}Zhejiang University, College of Electrical Engineering, 866 Yuhangtang Rd, Hangzhou, China, 310058,\IEEEauthorrefmark{2}HangZhou Dianzi University, China Institute of Computer Application Technology, Xiasha Higher Education Zone, Hangzhou, China, 310018}}

\markboth{Journal of \LaTeX\ Class Files,~Vol.~14, No.~8, August~2015}%
{Shell \MakeLowercase{\textit{et al.}}: Bare Demo of IEEEtran.cls for IEEE Transactions on Magnetics Journals}



% for Transactions on Magnetics papers, we must declare the abstract and
% index terms PRIOR to the title within the \IEEEtitleabstractindextext
% IEEEtran command as these need to go into the title area created by
\IEEEtitleabstractindextext{%
\begin{abstract}

Unsupervised learning methods for feature extraction are becoming more and more popular. We combine the popular contrastive learning method (prototypical contrastive learning) and the classic representation learning method (autoencoder) to design an unsupervised feature learning network for hyperspectral classification. Experiments have proved that our two proposed autoencoder networks have good feature learning capabilities by themselves, and the contrastive learning network we designed can better combine the features of the two to learn more representative features. As a result, our method surpasses other comparison methods in the hyperspectral classification experiments, including some supervised methods. Moreover, our method maintains a fast feature extraction speed than baseline methods. In addition, our method reduces the requirements for huge computing resources, separates feature extraction and contrastive learning, and allows more researchers to conduct research and experiments on unsupervised contrastive learning.

\end{abstract}

}

\maketitle

\IEEEdisplaynontitleabstractindextext

% \fi
\IEEEpeerreviewmaketitle

\section{Introduction}
\IEEEPARstart{W}{ith} the rapid development of deep learning, much progress has been made in the hyperspectral classification field. Recent years, many great supervised deep learning methods for hyperspectral classification were proposed, such as one-dimensional convolutional neural networks(1DCNN)\cite{1D-CNN}, three-dimensional convolutional neural networks(3DCNN)\cite{3DCNN}, deep recurrent neural networks (RNN)\cite{RNN}, deep feature fusion network(DFFN)\cite{DFFN}, spectral-spatial residual network(SSRN)\cite{SSRN}. Nowadays, it is widely admitted that the fusion of spatial information and spectral information is the best strategy to extract features for hyperspectral classification. Furthermore, deep learning models showed great power in finding the two kinds of information. With enough labeled samples, many supervised learning methods can effectively learn the spatial and spectral information simultaneously(e.g., SSRN) and update the models' parameters, thus getting good hyperspectral classification results. However, there are some disadvantages to supervised learning methods. Firstly, severe data dependence is the notorious disadvantage of supervised deep learning. Without enough labeled samples, many supervised deep learning methods can not work or only get bad results. Secondly, many supervised deep learning methods only rely on the labels to supervise the extraction of features. As a result, with the increase of labeled data, the models need to be retrained entirely, which seems expensive in the practical circumstances.

Researchers are also paying their attention to unsupervised learning methods to avoid the disadvantages of supervised deep learning. Unsupervised learning methods are mostly feature extraction algorithms, with a suitable feature extractor, we can use a simple classifier to get good classification results. To some extent, unsupervised learning methods are more convenient to be applied to practice. Speak of unsupervised learning, there are some classic methods widely applied to all kinds of fields, such as principal component analysis(PCA)\cite{PCA} and independent component analysis(ICA)\cite{ICA}. These traditional algorithms remain efficient in the pre-process or feature extraction field. 

Generally speaking, unsupervised learning aims to dig out information from data other than labels; there are two major ways to achieve the objective: representative learning and discriminative learning. There are two famous representative learning models: autoencoder\cite{AE} and generative adversarial network(GAN)\cite{GAN}. The two methods both use the self-similarity of data and then construct a model that can map the original data to some certain distributions. With the generated distributions, we can get distributions similar to the distributions of real samples. As a result, we can get forgery samples that are similar to real samples or get good feature extractors. Inspired by autoencoder, stacked sparse autoencoder(SSAE)\cite{SSAE} used autoencoder for sparse spectral feature learning and multiscale spatial feature learning, respectively and then fused these two kinds of feature for hyperspectral classification. Furthermore, GAN\cite{GANforC} also proved to be useful in extracting spectral-spatial features for hyperspectral classification. Except for these, 3D convolutional autoencoder(3DCAE)\cite{3DCAE}, semi-supervised GAN\cite{semi-GAN}, GAN-Assisted CapsNet(TripleGAN)\cite{triple-gan} and many other papers also originated from the representative learning. Generally, many researchers have acknowledged the utility of the representative learning methods in the remote sensing field.

On the other hand, discriminative learning is also an original idea in exploring the information of data. A typical way of using discriminative learning is the contrastive learning method. Unlike autoencoder and GAN, the contrastive learning method does not care about the details of the data. It cares about the difference between the different samples. In other words, contrastive learning aims to discriminate different samples and does not aim to find samples' exact distributions. By comparing different samples, a discriminator is trained and can be used for classification or other tasks. Because by finding the difference between two randomly selected samples, somehow the discriminator has learnt to extract features that are helpful for downstream tasks. Contrastive learning can be applied in supervised as well as unsupervised learning. For supervised contrastive learning, contrastive loss\cite{contrastiveloss} is proposed to optimize the discriminator. In the hyperspectral classification field, siamese convolutional neural networks\cite{scnn} adopted the original siamese network for scene classification. The end-to-end siamese convolutional neural network\cite{escnn} proposed a 1D siamese CNN  network for hyperspectral classification. Dual-path siamese convolutional neural network\cite{dualscnn} proposed a dual-path siamese network composed of 2D and 1D CNN for hyperspectral classification. Supervised learning has drawn many researchers' attention.

For unsupervised contrastive learning, there are no label information can be utilized at all. So a paradigm is widely used for unsupervised learning. Because contrastive learning requires the contrast between two samples, we use two kinds of data augmentation methods to transform a single sample into two transformed samples. In this way, we create a kind of label: the two samples originate from the same sample. Then with a proper structure and a well-designed objective function, we can train an encoder that can map the samples to latent encoding space, and the features from the encoding space can well represent the original samples. When the encoder acts as a feature extractor, many downstream tasks can be done. Much work has been done in the computer vision field, such as momentum contrast\cite{Moco} and a simple framework for contrastive learning\cite{simclr}. They proposed some original structures for unsupervised contrastive learning and got promising results in many visual datasets. Furthermore, prototypical contrastive learning(PCL)\cite{PCL} combined cluster algorithm with contrastive learning, got better performances than other methods, even outperformed some supervised methods. Generally, unsupervised contrastive learning has shown its great power in the computer vision field. 

Considering the similarity between hyperspectral images and normal images, we think it is possible to transfer the unsupervised contrastive learning methods to hyperspectral classification. However, as far as we know, few related methods have been reported in the hyperspectral classification field. There are some problems to address to transfer the unsupervised contrastive learning to the process of hyperspectral image. Firstly, the computer vision field's typical data augmentation methods seem not reasonable in the hyperspectral classification field. For example, color distortion to the normal images will not change objects' spatial features, so it is an acceptable transformation. When distorting hyperspectral images' spectral value, the spectral information is destroyed, and the spectral information is essential for hyperspectral classification, so this is not an acceptable transformation for hyperspectral images. Secondly, unsupervised contrastive learning for normal images demands large batch size and lots of GPUs, if it is directly applied to hyperspectral images, the requirement for more computing resources will be too expensive. 

To apply the unsupervised contrastive learning to hyperspectral classification, we addressed the two problems and proposed ContrastNet, a deep learning model for unsupervised feature learning of hyperspectral images. Our innovations can be concluded as follows:
\begin{enumerate}[1)]
\item  We introduced discriminative learning methods into hyperspectral images. Moreover, we applied the prototypical contrastive learning to unsupervised feature learning of hyperspectral classification.
\item We designed an excellent pipeline to process the hyperspectral images, which combines two kinds of unsupervised learning methods (representative learning and contrastive learning). Furthermore, the pipeline only demands a single GPU, which means it is more affordable than the original prototypical contrastive learning method.
\item Enlighted by the autoencoder structure, we designed two autoencoder-based modules: adversarial autoencoder module(AAE module), variational autoencoder module(VAE module). The two modules can work alone for feature extraction. When using the extracted features for hyperspectral classification, they outperformed many baselines, even supervised methods.
\end{enumerate}

\section{Related Work}
\subsection{Variational Autoencoder}
Autoencoder structure(AE)\cite{AE} is a widely used unsupervised feature extraction model. The standard autoencoder is composed of an encoder and a decoder. The encoder encodes the input images to features with fixed shape. Moreover, the features are called latent code. The decoder decodes the latent code to original input images. The autoencoder's objective function is usually reconstruction loss(e.g., Euclidean distances between output and input). Furthermore, the whole model is optimized by the backpropagation algorithm. 

However, original AE does not restrict the distribution of latent code. Thus variational autoencoder(VAE)\cite{VAE} is proposed to optimize the distribution of latent code. VAE used KL-divergence and the reparameterization trick to restrict latent code distribution so that randomly sampled value from a normal distribution can be decoded to an image with similar distribution to the training samples. The latent code generated by VAE performs better in the classification task, so VAE is widely used for unsupervised feature extraction.
\subsection{Adversarial Autoencoder}
Adversarial autoencoder(AAE)\cite{AAE} is another AE-based structure which adopts adversarial restriction on the latent code. Unlike VAE, AAE does not use KL-divergence to measure the distance between latent code distribution and normal distribution. AAE chooses to use the idea of adversarial to restrict the distribution of latent code. By training a generative adversarial network(GAN)\cite{GAN} with the latent code and randomly selected samples from a normal distribution, the distribution of latent code will become an expected normal distribution. As a result, the latent code generated by AAE will be great for classification, and the distributions of it will be similar but different from that generated by VAE.
\subsection{Prototypical Contrast Learning}
Usual unsupervised contrastive learning methods aim to find an embedding space where every sample can be discriminated among other instances in the whole dataset, such as momentum contrast(MoCo)\cite{Moco}. But this objective may be too hard to achieve without enough computing resources or sufficient iteration. Although this kind of distribution may be beneficial in some downstream tasks, it is not very fit for classification. Excellent distribution for classification demands for clear clusters composed of the data in the same classes. And typical unsupervised contrastive learning can't satisfy this demand.

Prototypical contrastive learning(PCL)\cite{PCL} is an unsupervised representation learning method based on contrastive learning. Unlike most contrastive learning methods, PCL introduces prototypes as latent variables to help find the maximum-likelihood estimation of the network parameters in an expectation-maximization framework. PCL uses ProtoNCE loss, a generalized version of the InfoNCE\cite{InfoNCE} loss for contrastive learning by encouraging representations to be closer to their assigned prototypes. The experiments show that PCL can get better performances than other baselines in the classification tasks, so we choose PCL as our method's base structure.

\section{Proposed Method}
\subsection{Motivation}
PCL showed great power of unsupervised learning in processing normal images. However, as mentioned above, it is hard to apply PCL to hyperspectral images directly. So we must make some adjustments to PCL in order to apply it to hyperspectral classification. With the blossom of AE-based methods in the hyperspectral images field, we find that the encoder in an autoencoder structure can be seen as a transformation, just like other data augmentation methods. They both map the original images from the original distribution to another distribution. The difference is, data augmentations usually do not change the shape of images, while encoder usually encodes the images to vectors with less dimensions. Another essential truth is that data augmentation methods usually keep most information of the original images to make them able to be rightly classified. Moreover, the encoder in a trained autoencoder also can keep most information in the original images.

As a result, we find it reasonable to treat the encoder as a transformation function that plays the role of data augmentation. Then the question is to find two encoders with different distributions, just like two kinds of data augmentations. So we choose designed AAE and VAE modules to act as two kinds of data augmentation because AAE and VAE can generate relatively fixed and slightly different distributions.

In this way, we can not only apply PCL to hyperspectral images but also lower the demand for deep networks. By encoding the images to shape fixed vectors(e.g., 1024-d), we can assign the feature extraction and contrastive learning into two parts, and each part will not be too complicated. Through this method, we make contrastive learning methods affordable for those who have limited computing resources.

\subsection{VAE Module}
\begin{figure*}[!t]
\centering
\includegraphics[width=6.5in]{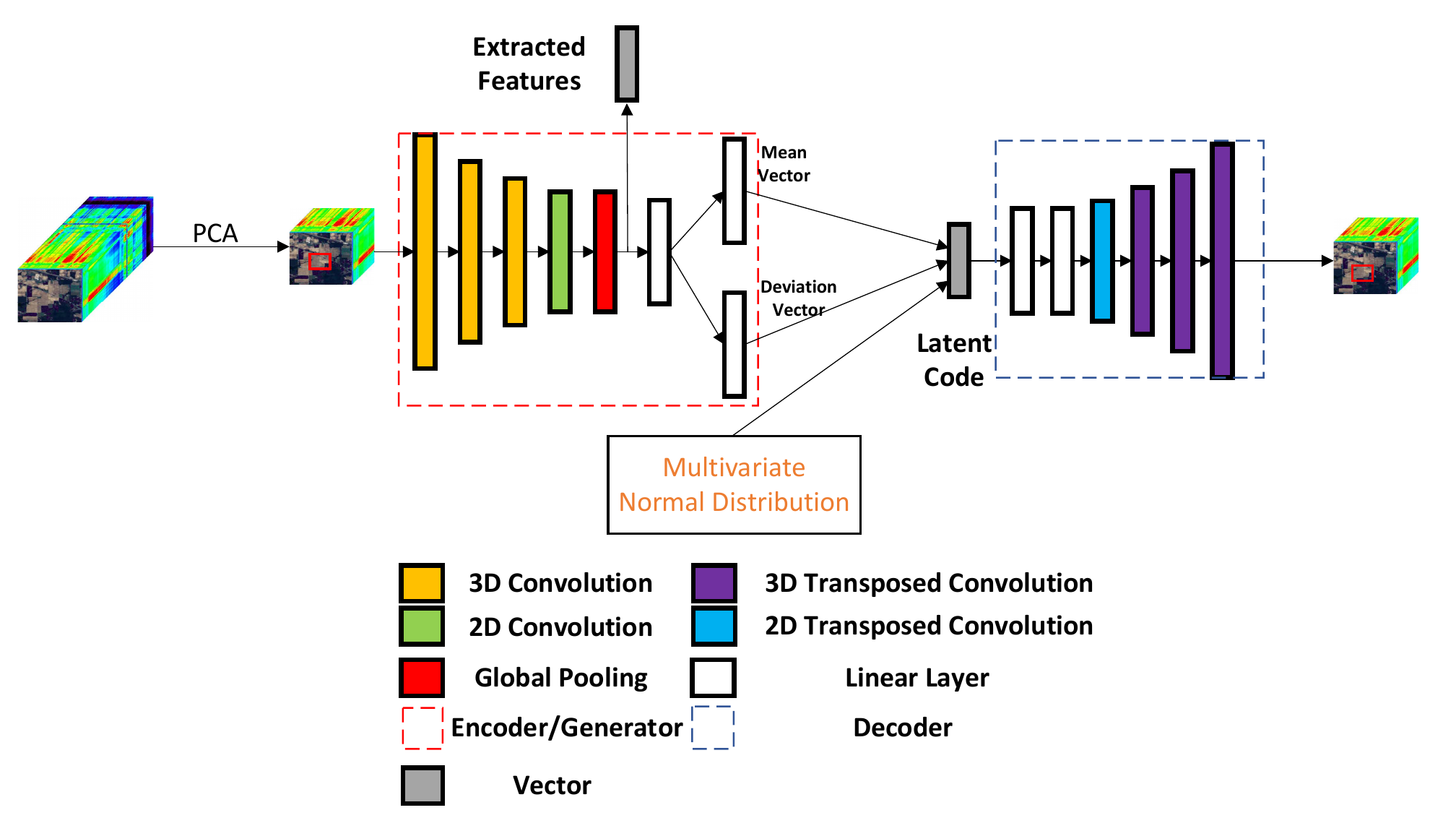}
\caption{Structure of VAE module. Every (transposed) convolution comprises a (transposed) convolutional layer and a batch normalization layer. Every linear layer means a linear layer and a ReLU layer.}
\label{fig:VAE}
\end{figure*}

To process the hyperspectral images, we design a VAE network to extract features from the original hyperspectral images. The structure of the VAE module is shown in Fig.\ref{fig:VAE}. As the figure shows, we first use PCA preprocess the whole hyperspectral image to cut down the channels. Then sliding windows are adopted to split the image into small patches with the shape $S \times S \times K$. S is the size of sliding windows, and K is the channel of the preprocessed hyperspectral image. The structure of the encoder network is enlighted by the idea of hybrid convolution in HybridSN\cite{HybridSN}. We used several 3D convolutional layers and 2D convolutional layers to extract the input patches' spectral and spatial information. We also used several 3D transposed convolutional layers and 2D transposed convolutional layers to build the decoder network.

Furthermore, to avoid the risk of overfitting, we adopted batch normalization\cite{bn} after every convolutional layer. Every linear layer is combined with a  rectified linear unit(ReLU) activation layer. We use a global pooling layer to fix the shape of the feature map, thus fix the shape of extracted features(e.g., 1024-d) after reshaping operation.

According to the original VAE, we let the encoder output two variables $\mu$ and $\sigma$. Then the latent code can be computed by Eq. \ref{eqa:vae_code}. $z$ is the latent code, and $\varepsilon$ is a random value drawn from the normal distribution. The loss function for VAE network can be divided into two parts: $\mathcal{L}_{\mu, \sigma^{2}}$ and $\mathcal{L}_{recon}$. As Eq. \ref{eqa:vae_ll} shows, $N$ is the number of samples, $\mathcal{L}_{\mu, \sigma^{2}}$ computes the KL divergence between latent code and normal distribution. Moreover, Eq. \ref{eq:recon} shows the detail of reconstruction loss $\mathcal{L}_{recon}$, $N$ is the number of samples, the input image patch is $I^{x, y, z}$ with the shape of $s \times s \times k$, and $\hat{I}^{x, y, z}$ is the output image patch of the decoder. The whole loss function can be represented by Eq. \ref{eqa:vaeloss}.

\begin{equation}
\label{eqa:vae_code}
{
z=\mu+\varepsilon \times \sigma
}
\end{equation}

\begin{equation}
\label{eqa:vae_ll}
{
\mathcal{L}_{\mu, \sigma^{2}}=\frac{1}{2} \sum_{i=1}^{N}\left(\mu_{(i)}^{2}+\sigma_{(i)}^{2}-\log \sigma_{(i)}^{2}-1\right)}
\end{equation}

\begin{equation}
\label{eq:recon}
\mathcal{L}_{recon}=\sum_{i=1}^{N}\left( \frac{1}{s \times s \times k} \sum_{x=0}^{s-1} \sum_{y=0}^{s-1} \sum_{z=0}^{k-1}\left(I^{x, y, z}-\hat{I}^{x, y, z}\right)^{2}\right)
\end{equation}

\begin{equation}
\label{eqa:vaeloss}
\mathcal{L}=\mathcal{L}_{\mu, \sigma^{2}}+\mathcal{L}_{recon}
\end{equation}

When we finished the VAE module training, we can map all the patches to VAE features. Thus we can use these VAE features to build the training dataset for ContrastNet.
\subsection{AAE Module}
\begin{figure*}[!t]
\centering
\includegraphics[width=6.5in]{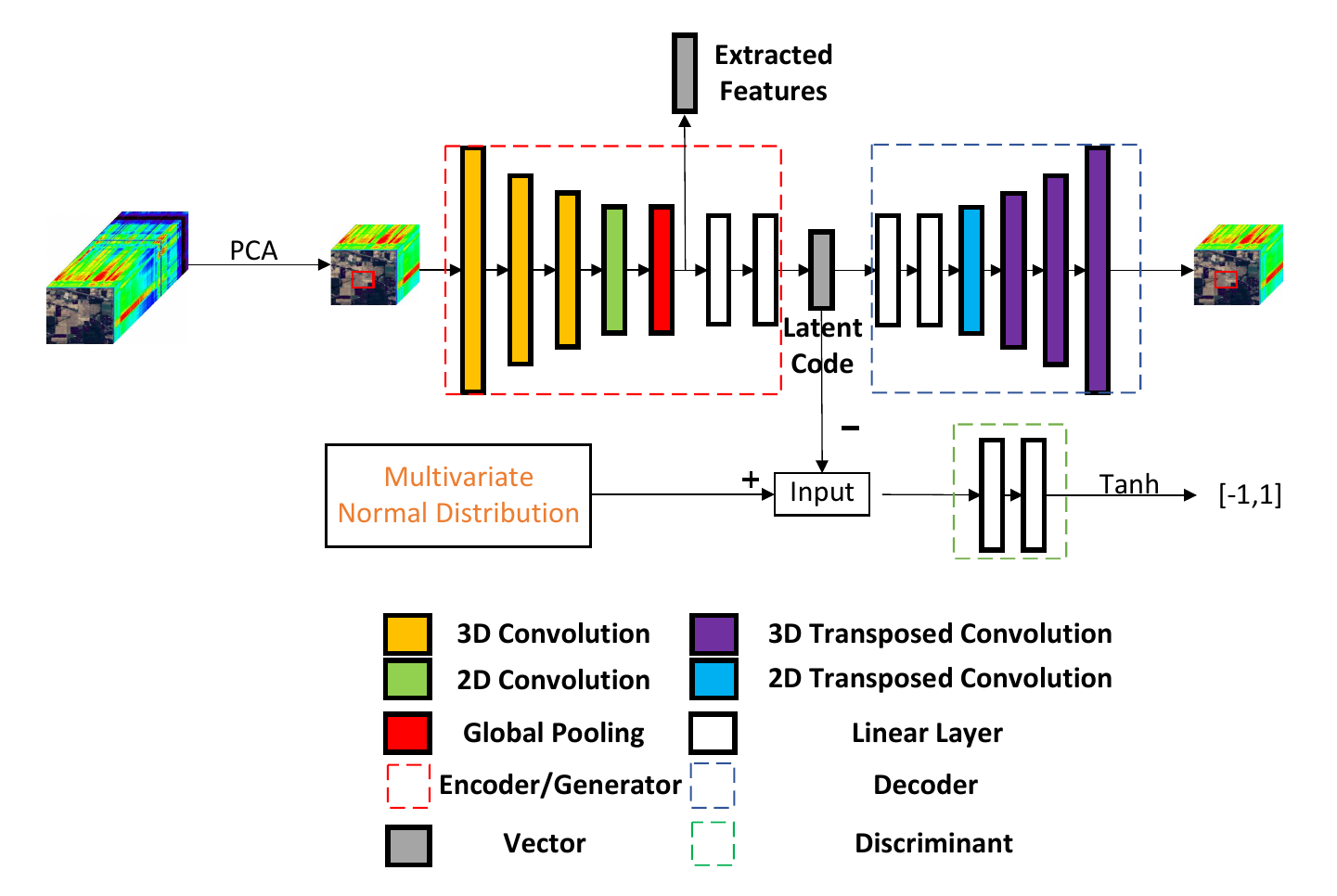}

\caption{Structure of AAE module. Every (transposed) convolution comprises a (transposed) convolutional layer and a batch normalization layer. Every linear layer means a linear layer and a ReLU layer.}
\label{fig:AAE}
\end{figure*}

Similar to the VAE module, the whole hyperspectral image is preprocessed by PCA in the AAE module. The structure of the AAE network is shown in Fig.\ref{fig:AAE}. The encoder and the decoder in the AAE module are the same with that in the VAE module except for the linear layers before the latent code. The encoder in the AAE module directly outputs the latent code and acts as a GAN generator.

According to the original AAE, the latent code is the output of the generator and the discriminant input. Another input of the discriminant is a randomly drawn variable from the normal distribution. The training of the AAE can be divided into two phases: reconstruction phase and regularization phase. In the reconstruction phase, we use reconstruction loss to optimize the parameters in the network. Reconstruction loss is shown in Eq. \ref{eq:recon2}, it is the same as the reconstruction loss in the VAE module. $\mathcal{L}_{recon}$ is reconstruction loss, $N$ is the number of samples, $I^{x, y, z}$ represents the input image patch, $s \times s \times k$ represents the shape of the input patch and $\hat{I}^{x, y, z}$ means reconstructed image patch.

In the regularization phase, the discriminator and the encoder are optimized. At first, we optimize the discriminator and then the generator(encoder). We adopt Wasserstein GAN(WGAN)\cite{WGAN} loss as our optimization function for the discriminator and the generator. Discriminant loss and generator loss are shown in Eq. \ref{eqa:LD} and \ref{eqa:LG}, where  $\mathcal{L}_D$ represents the loss of the discriminant, and $\mathcal{L}_G$ represents the loss of the generator. $P_{g}$ is the distribution of the generated samples(latent code), and $P_{r}$ is the distribution of the real samples(samples from a normal distribution). x represents random samples from $P_{g}$ and $P_{r}$.

\begin{equation}
\label{eq:recon2}
\mathcal{L}_{recon}=\sum_{i=1}^{N}\left( \frac{1}{s \times s \times k} \sum_{x=0}^{s-1} \sum_{y=0}^{s-1} \sum_{z=0}^{k-1}\left(I^{x, y, z}-\hat{I}^{x, y, z}\right)^{2}\right)
\end{equation}
\begin{equation}
\label{eqa:LD}{
\mathcal{L}_D=\mathbb{E}_{x \sim P_{g}}\left[\mathbf{D}(x)\right]-\mathbb{E}_{x \sim P_{r}}\left[\mathbf{D}(x)\right]
}
\end{equation}

\begin{equation}
\label{eqa:LG}
{
\mathcal{L}_G=-\mathbb{E}_{x \sim P_{g}}\left[\mathbf{D}(x)\right]
}
\end{equation}

Similarly, when we finished the training of the AAE module, we can map all the patches to AAE features. Thus we can use these AAE features to build the training dataset for ContrastNet.

\subsection{ContrastNet}
Before we introduce the ContrastNet, the specific description of typical unsupervised contrast learning should be clarified. Let the traing set $X=\{{x_1,x_2,...,x_n}\}$ have $n$ images. And our goal is to find an embedding function $f_\theta$(realized via a deep neural network) that maps $X$ to another space $V=\{{v_1,v_2,...,v_n}\}$ where $v_i=f_\theta(x_i)$, and $v_i$ can describe $x_i$ best. Usually the objective is achieved by the InfoNCE loss function\cite{cpc} as Eq. \ref{eq:infoNCE} shows. $v'_i$ is a positive embedding for instance $i$, and $v'_j$ includes one positive embedding and r negative embeddings for other instances, and $\tau$ is a temperature hyper-parameter. In MoCo\cite{Moco}, these embeddings are obtained by feeding $x_i$ to a momentum encoder parametrized by $\theta'$, $v'_i=f_{\theta'}(x_i)$, where $\theta'$ is a moving average of $\theta$.

When minimizing the Eq. \ref{eq:infoNCE}, the distance between $v_i$ and $v'_i$ are becoming close, and the distance between $v_i$ and $r$ negative embeddings are becoming far. However, original InfoNCE loss uses a fixed $\tau$, which means the same concentration on a certain scale. Moreover, it computes the distance between $v_i$ and all kinds of negative embeddings, some of which are not representative.

As a result, prototypical contrastive learning proposed a new loss function named ProtoNCE. As Eq. \ref{eq:Proto} and Eq. \ref{eq:ProtoNCE} show, based on InfoNCE, ProtoNCE contains a second part $\mathcal{L}_{Proto}$. It use prototypes $c$ instead of $v'$, and replace the fixed temperature $\tau$ with a per-prototype concentration estimation $\phi$. 

\begin{equation}
\label{eq:infoNCE}{
\mathcal{L}_{\mathrm{InfoNCE}}=\sum_{i=1}^{n}-\log \frac{\exp \left(v_{i} \cdot v_{i}^{\prime} / \tau\right)}{\sum_{j=0}^{r} \exp \left(v_{i} \cdot v_{j}^{\prime} / \tau\right)}
}
\end{equation}

\begin{equation}
\label{eq:Proto}
\mathcal{L}_{Proto}= \sum_{i=1}^{n}-\left(\frac{1}{M} \sum_{m=1}^{M} \log \frac{\exp \left(v_{i} \cdot c_{s}^{m} / \phi_{s}^{m}\right)}{\sum_{j=0}^{r} \exp \left(v_{i} \cdot c_{j}^{m} / \phi_{j}^{m}\right)}\right)
\end{equation}

\begin{equation}
\label{eq:ProtoNCE}
\mathcal{L}_{ProtoNCE}= \mathcal{L}_{Proto} +\mathcal{L}_{\mathrm{InfoNCE}} 
\end{equation}

The detailed algorithm of prototypical contrastive learning can be found in Algorithm \ref{alg:PCL}. There are two encoders in the PCL algorithm, one encoder maps $x_i$ to $v_i$, and the momentum encoder maps $x_i$ to $v'_i$. Practically, we do not use the same $x_i$ to get $v_i$ and $v'_i$, two images with proper data augmentation will be the input of two encoders for better performance. The first encoder parameters can be represented by $\theta$, and the parameters in the momentum encoder can be represented by $\theta'$ which meets Eq. \ref{eq:theta}. In algorithm \ref{alg:PCL}, $k-means()$\cite{faiss} is the GPU implementation of a widely used clustering algorithm. And $C^m$ is the sets of $k_m$ prototypes(center of clusters). $Concentration()$ is the function of concentration estimation $\phi$, as Eq. \ref{eq:phi} shows, $\phi$ is calculated by the momentum features $\{v'_z\}^Z_{z=1}$ that are within the same clusters as a prototype $c$, $\alpha$ is a smooth parameter to ensure that small clusters do not have overly-large $\phi$, and the typical value is 10. $SGD()$\cite{sgd} is a kind of optimization function to update the parameters in the encoder.

\begin{algorithm}[!t]
\renewcommand{\algorithmicrequire}{\textbf{Input:}}
\renewcommand{\algorithmicensure}{\textbf{Output:}}
\caption{Prototypical Contrastive Learning.}
\label{alg:PCL}
\begin{algorithmic}[1]
\Require encoder $f_\theta$, training dataset $X$, number of clusters $K=\{k_m\}^M_{m=1}$
\State $\theta'=\theta$
\While{not MaxEpoch} 

\State $V'=f_\theta'(X)$
	\For{$m = 1$ \textbf{to} $M$}
	\State $C^m=k-means(V',k_m)$
	\State $\phi^m=Concentration(C^m,V')$
	\EndFor
	\For{$x$ \textbf{in} $Dataloader(X)$}
	\State $v=f_\theta(x),v'=f'_\theta(x)$
	\State $\mathcal{L}_{ProtoNCE}(v,v',\{C^m\}^M_{m=1},\{\phi^m\}^M_{m=1})$
	\State $\theta=SGD(\mathcal{L}_{ProtoNCE},\theta)$
	\State $\theta'=0.999*\theta'+0.001*\theta$
	\EndFor
\EndWhile 
\end{algorithmic}
\end{algorithm}

\begin{equation}
\label{eq:theta}
\theta'=0.999*\theta'+0.001*\theta
\end{equation}

\begin{equation}
\label{eq:phi}
\phi=\frac{\sum_{z=1}^{Z}\left\|v_{z}^{\prime}-c\right\|_{2}}{Z \log (Z+\alpha)}
\end{equation}

The proposed ContrastNet is similar to the original PCL algorithm, but there are several differences. Firstly, ContrastNet uses AAE and VAE features as inputs, not two perturbed images. Secondly, the structure of ContrastNet is more straightforward and specialized in the "contrastive" part, not the "encoding" part. Thirdly, we adopt the projection head trick in sim-CLR\cite{simclr} to improve the performance.
\begin{figure*}[!t]
\centering
\includegraphics[width=6.5in]{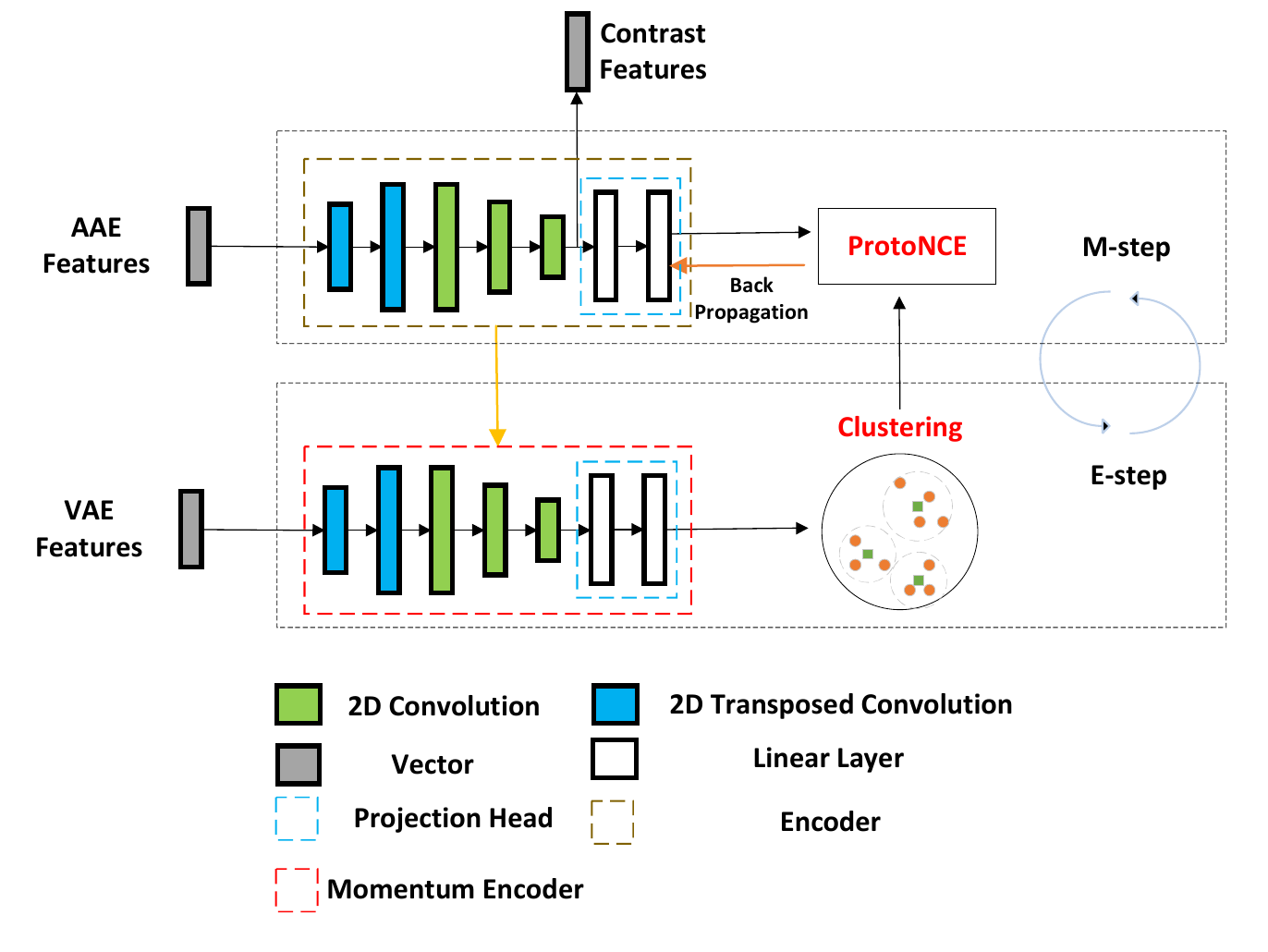}

\caption{Structure of ContrastNet. Every (transposed) convolution comprises a (transposed) convolutional layer and a batch normalization layer. Every linear layer means a linear layer and a ReLU layer. Clustering means k-means algorithm, and ProtoNCE is $\mathcal{L}_{ProtoNCE}$ in Eq. \ref{eq:ProtoNCE}. M-step and E-step are the phases of Expectation-Maximization algorithm. }
\label{fig:ContrastNet}
\end{figure*}

Fig. \ref{fig:ContrastNet} is an overview of the ContrastNet. Expectation-Maximization (EM) algorithm is adopted to train the ContrastNet. In the E-step, VAE features are fed into the momentum encoder, then $C^m$ and $\phi^m$ are calculated. In the M-step, $\mathcal{L}_{ProtoNCE}$ is calculated based on the updated features and variables in the E-step, then the two encoders are updated via backpropagation.
Similarly, all convolutions in the ContrastNet is composed of a convolutional layer and a batch normalization layer, and every linear layer is combined with a ReLU layer. The 1024-d AAE and VAE features will be reshaped to $4 \times 4 \times 64$ feature maps first, in this way, the spatial information extracted can be preserved. And then, the feature maps are fed into the ContrastNet. The structure of ContrastNet is light and straightforward because the previous process has achieved the feature extraction. The ContrastNet only needs to pay attention to the contrastive learning. We adopted the projection head structure in the ContrastNet, that is, we use the output of the projection head for training, and use the features before the head for testing. In this way, the extracted contrast features will contain more original information. In the testing phase, we need to extract contrast features in Fig. \ref{fig:ContrastNet} from the primary encoder, which can save much testing time for the ContrastNet.

\section{Experiments and Analysis}
\subsection{Datasets}

Experiments are conducted using three publicly available HSI datasets: Indian Pines(IP), University of Pavia(PU), and Salinas Scene(SA). The IP dataset was gathered using the Airborne Visible / Infrared Imaging Spectrometer (AVIRIS)  hyperspectral sensor. It consisted of 145$\times$145 pixels and 224 spectral reflectance bands in the wavelength range of $0.4\sim2.5 \times10^{-6}$ meters. After removing bands covering the region of water absorption, 200 bands were left, and the ground truth was composed of 16 classes. The PU dataset was a hyperspectral image with 610$\times$340 pixels and 103 spectral bands in the wavelength range of $0.43\sim0.86 \times10^{-6}$ meters. The ground truth contained nine classes. The SA dataset was an image with 512$\times$217 pixels and 224 spectral bands. This dataset was in the wavelength range of $0.36\sim2.5 \times10^{-6}$ meters. Twenty water-absorbing bands were discarded from the SA dataset, so it was a dataset with 204 spectral bands and 16 landcover classes in the ground truth.

\subsection{Experiments Details}
All experiments were implemented on a PC with 16GB RAM and a GTX 1080 GPU. The coding environment was pytorch\cite{pytorch}. The shape of patches is $27 \times 27 \times 15$ in PU and SA dataset, and $27 \times 27 \times 30$ in the IP dataset. When we use $27 \times 27 \times 15$ patches, our networks' details are shown as follows.

Tab. \ref{tab:AAE_enc} and Tab. \ref{tab:AAE_dec} show the structure of AAE. The structure of the discriminant is simple, so it is omitted. The latent code dimension is 128 and extracted AAE features are flattened outputs of the pooling layer in the encoder(1024-d). In the training phase, we use two Adam\cite{adam} optimizers to optimize the encoder and the decoder, and two SGD\cite{sgd} optimizers to optimize the generator and the discriminant. Two Adam optimizers are set with the learning rate of 0.001, and weight decay is set to 0.0005. The optimizer for the generator is set with a 0.0001 learning rate and no weight decay. The optimizer for the discriminant is set with a 0.00005 learning rate and no weight decay. The batch size is set to 128. We train the AAE module for 20 epochs, and only save the parameters of the encoder.
\begin{table}[!t]
\caption{Structure of AAE Encoder. The shape of data is defined in pytorch style. -1 means batchsize in the shape array.}
\label{tab:AAE_enc}

\centering
\begin{tabular}{|c|c|}
\hline
\multicolumn{2}{|c|}{AAE Encoder}       \\ \hline
Layer(type)       & Output Shape        \\ \hline
input             & {[}-1,1,15,27,27{]} \\ \hline
Conv3d            & {[}-1,8,9,25,25{]}  \\ \hline
BatchNorm3d       & {[}-1,8,9,25,25{]}  \\ \hline
ReLU              & {[}-1,8,9,25,25{]}  \\ \hline
Conv3d            & {[}-1,16,5,23,23{]} \\ \hline
BatchNorm3d       & {[}-1,16,5,23,23{]} \\ \hline
ReLU              & {[}-1,16,5,23,23{]} \\ \hline
Conv3d            & {[}-1,32,3,21,21{]} \\ \hline
BatchNorm3d       & {[}-1,32,3,21,21{]} \\ \hline
ReLU              & {[}-1,32,3,21,21{]} \\ \hline
Conv2d            & {[}-1,64,19,19{]}   \\ \hline
BatchNorm2d       & {[}-1,64,19,19{]}   \\ \hline
ReLU              & {[}-1,64,19,19{]}   \\ \hline
AdaptiveAvgPool2d & {[}-1,64,4,4{]}     \\ \hline
Linear            & {[}-1,512{]}        \\ \hline
ReLU              & {[}-1,512{]}        \\ \hline
Linear            & {[}-1,128{]}        \\ \hline
\end{tabular}
\end{table}

\begin{table}[]
\caption{Structure of AAE Decoder. The shape of data is defined in pytorch style. -1 means batchsize in the shape array.}
\label{tab:AAE_dec}
\centering
\begin{tabular}{|c|c|}
\hline
\multicolumn{2}{|c|}{AAE Decoder}      \\ \hline
Layer(type)      & Output Shape        \\ \hline
input            & {[}-1,128{]}        \\ \hline
Linear           & {[}-1,256{]}        \\ \hline
ReLU             & {[}-1,256{]}        \\ \hline
Linear           & {[}-1,23104{]}      \\ \hline
ReLU             & {[}-1,23104{]}      \\ \hline
ConvTransposed2d & {[}-1,96,21,21{]}   \\ \hline
BatchNorm2d      & {[}-1,96,21,21{]}   \\ \hline
ReLU             & {[}-1,96,21,21{]}   \\ \hline
ConvTransposed3d & {[}-1,16,5,23,23{]} \\ \hline
BatchNorm3d      & {[}-1,16,5,23,23{]} \\ \hline
ReLU             & {[}-1,16,5,23,23{]} \\ \hline
ConvTransposed3d & {[}-1,8,9,25,25{]}  \\ \hline
BatchNorm3d      & {[}-1,8,9,25,25{]}  \\ \hline
ReLU             & {[}-1,8,9,25,25{]}  \\ \hline
ConvTransposed3d & {[}-1,1,15,27,27{]} \\ \hline
BatchNorm3d      & {[}-1,1,15,27,27{]} \\ \hline
\end{tabular}
\end{table}

Tab. \ref{tab:VAE_enc} and Tab. \ref{tab:VAE_dec} show the structure of VAE. The dimension of latent code is also 128 and extracted VAE features are flattened outputs of the pooling layer in the encoder(1024-d) too. In the training phase, we use two Adam\cite{adam} optimizers to optimize the encoder and the decoder. Two Adam optimizers are set with the learning rate of 0.001, and weight decay is set to 0.0005. The two outputs of VAE encoder are used to compute the latent code. Then the latent code is fed into the decoder. The batch size is set to 128. We train the VAE module for 30 epochs, and only save the parameters of the encoder.

\begin{table}[]
\caption{Structure of VAE Encoder. The shape of data is defined in pytorch style. -1 means batchsize in the shape array.}
\label{tab:VAE_enc}
\centering
\begin{tabular}{|c|c|}
\hline
\multicolumn{2}{|c|}{VAE Encoder}             \\ \hline
input             & {[}-1,1,15,27,27{]}       \\ \hline
Conv3d            & {[}-1,8,9,25,25{]}        \\ \hline
BatchNorm3d       & {[}-1,8,9,25,25{]}        \\ \hline
ReLU              & {[}-1,8,9,25,25{]}        \\ \hline
Conv3d            & {[}-1,16,5,23,23{]}       \\ \hline
BatchNorm3d       & {[}-1,16,5,23,23{]}       \\ \hline
ReLU              & {[}-1,16,5,23,23{]}       \\ \hline
Conv3d            & {[}-1,32,3,21,21{]}       \\ \hline
BatchNorm3d       & {[}-1,32,3,21,21{]}       \\ \hline
ReLU              & {[}-1,32,3,21,21{]}       \\ \hline
Conv2d            & {[}-1,64,19,19{]}         \\ \hline
BatchNorm2d       & {[}-1,64,19,19{]}         \\ \hline
ReLU              & {[}-1,64,19,19{]}         \\ \hline
AdaptiveAvgPool2d & {[}-1,64,4,4{]}           \\ \hline
Linear            & {[}-1,512{]}              \\ \hline
ReLU              & {[}-1,512{]}              \\ \hline
Linear/Linear     & {[}-1,128{]}/{[}-1,128{]} \\ \hline
\end{tabular}
\end{table}

\begin{table}[]
\caption{Structure of VAE Decoder. The shape of data is defined in pytorch style. -1 means batchsize in the shape array.}
\label{tab:VAE_dec}
\centering
\begin{tabular}{|c|c|}
\hline
\multicolumn{2}{|c|}{VAE Decoder}      \\ \hline
Layer(type)      & Output Shape        \\ \hline
input            & {[}-1,128{]}        \\ \hline
Linear           & {[}-1,256{]}        \\ \hline
ReLU             & {[}-1,256{]}        \\ \hline
Linear           & {[}-1,23104{]}      \\ \hline
ReLU             & {[}-1,23104{]}      \\ \hline
ConvTransposed2d & {[}-1,96,21,21{]}   \\ \hline
BatchNorm2d      & {[}-1,96,21,21{]}   \\ \hline
ReLU             & {[}-1,96,21,21{]}   \\ \hline
ConvTransposed3d & {[}-1,16,5,23,23{]} \\ \hline
BatchNorm3d      & {[}-1,16,5,23,23{]} \\ \hline
ReLU             & {[}-1,16,5,23,23{]} \\ \hline
ConvTransposed3d & {[}-1,8,9,25,25{]}  \\ \hline
BatchNorm3d      & {[}-1,8,9,25,25{]}  \\ \hline
ReLU             & {[}-1,8,9,25,25{]}  \\ \hline
ConvTransposed3d & {[}-1,1,15,27,27{]} \\ \hline
BatchNorm3d      & {[}-1,1,15,27,27{]} \\ \hline
\end{tabular}
\end{table}

The two encoders in ContrastNet have the same structure, which is shown in Tab. \ref{tab:Contrast_enc}. Two inputs are 1024-d vectors, and the dimension of contrast features and projected features are both 128-d. We train the ContrastNet with an SGD optimizer, and the learning rate is 0.003, weight decay is set to 0.001. Number of negative samples $r=640$, and batch size is 128, $\tau$ in Eq. \ref{eq:infoNCE} is 0.01, number of clusters $K=[1000,1500,2500]$. According to the original PCL method, we only train the network with Eq. \ref{eq:infoNCE} for 30 epochs, then train the network with Eq. \ref{eq:ProtoNCE} for 170 epochs. The learning rate will be multiplied with 0.1 when the trained epoch is more than 120 and 160.

\begin{table}[]
\caption{Structure of ContrastNet Encoder. The shape of data is defined in pytorch style. -1 means batchsize in the shape array.}
\label{tab:Contrast_enc}
\centering
\begin{tabular}{|c|c|}
\hline
\multicolumn{2}{|c|}{ContrastNet Encoder} \\ \hline
Layer(type)         & Output Shape        \\ \hline
input               & {[}-1,1024{]}       \\ \hline
ConvTransposed2d    & {[}-1,64,6,6{]}     \\ \hline
BatchNorm2d         & {[}-1,64,6,6{]}     \\ \hline
ReLU                & {[}-1,64,6,6{]}     \\ \hline
ConvTransposed2d    & {[}-1,64,8,8{]}     \\ \hline
BatchNorm2d         & {[}-1,64,8,8{]}     \\ \hline
ReLU                & {[}-1,64,8,8{]}     \\ \hline
Conv2d              & {[}-1,128,6,6{]}    \\ \hline
BatchNorm2d         & {[}-1,128,6,6{]}    \\ \hline
ReLU                & {[}-1,128,6,6{]}    \\ \hline
Conv2d              & {[}-1,64,4,4{]}     \\ \hline
BatchNorm2d         & {[}-1,64,4,4{]}     \\ \hline
ReLU                & {[}-1,64,4,4{]}     \\ \hline
Conv2d              & {[}-1,32,2,2{]}     \\ \hline
ReLU                & {[}-1,32,2,2{]}     \\ \hline
Linear              & {[}-1,128{]}        \\ \hline
ReLU                & {[}-1,128{]}        \\ \hline
Linear              & {[}-1,128{]}        \\ \hline
\end{tabular}
\end{table}
\subsection{Results}
We conducted classification experiments in the three datasets. Eleven other methods were adopted as baselines to compare with our methods. They are linear discriminant analysis(LDA)\cite{LDA}, locality-preserving dimensionality reduction(LFDA)\cite{LFDA}, sparse graph-based discriminant analysis(SGDA)\cite{SGDA}, sparse and low-rank graph for discriminant analysis(SLGDA)\cite{SLGDA}, deep convolutional neural networks(1D-CNN)\cite{1D-CNN}, supervised deep feature extraction(S-CNN)\cite{S-CNN}, tensor principal component analysis(TPCA)\cite{TPCA}, stacked sparse autoencoder(SSAE)\cite{SSAE}, unsupervised deep feature extraction(EPLS)\cite{EPLS}, and 3D convolutional autoencoder(3DCAE)\cite{3DCAE}. To illustrate the great feature learning ability of ContrastNet, we used SVM to classify the features we learned by ContrastNet. Moreover, features learned by AAE and VAE were also used to compare with the ContrastNet. 10\% samples of each class are randomly selected for training SVM in the IP and PU datasets, and 5\% samples of each class are used for training in the SA dataset, the rest in each dataset is used for testing. The quantitative results are evaluated by average accuracy (AA) and overall accuracy (OA). The results in the Indian Pines dataset are shown in Tab. \ref{tab:IP}. It is clear that the structures we proposed performed best in many classes. AAE got the best results in five classes while ContrastNet got the best results in six classes, and VAE also performed best in 3 classes.

To some extent, AAE and VAE are two better feature extractors than 3DCAE in terms of AA. However, ContrastNet, which learns information from AAE and VAE, showed a significant increase in OA. It can be evidence indicating that ContrastNet can extract some core information from two distributions by contrast learning. As a result, it got the highest OA. However, ContrastNet did not perform well enough in AA in the Indian Pines dataset. We guess the reason for it may be the imbalance of the number of samples in the Indian Pines dataset. Classes with large numbers dominated the location of clusters. Thus the ContrastNet tended to perform well in these classes, which decreased accuracy in other classes.

\begin{table*}[htpb]
\caption{Comparison in the Indian Pines dataset. 10\% labeled samples are used for training, and the rest are used for testing. AA menas average accuracy and OA means overall accuracy. The best values in each row are in bold. And the methods in bold are structures proposed in this paper. Some of the data in the table is quoted from \cite{3DCAE}. }
\label{tab:IP}
\begin{tabular}{c|cccccc|cccccccc}
\hline
\multirow{2}{*}{Class} & \multicolumn{6}{c|}{Supervised Feature Extraction}      & \multicolumn{8}{c}{Unsupervised Feature Extraction}                                                                    \\ \cline{2-15} 
                       & LDA   & LFDA  & SGDA  & SLGDA          & 1D-CNN & S-CNN & PCA   & TPCA            & SSAE            & EPLS  & 3DCAE          & \textbf{AAE}             & \textbf{VAE}             & \textbf{ContrastNet}       \\ \hline
1                      & 58.54 & 29.63 & 42.59 & 39.62          & 43.33  & 83.33 & 39.02 & 60.97           & 56.25           & 58.72 & 90.48          & \textbf{100.00} & \textbf{100.00} & 85.37          \\ \hline
2                      & 69.88 & 75.59 & 80.89 & 85.56          & 73.13  & 81.41 & 72.30 & 87.00           & 69.58           & 59.91 & 92.49          & 81.63           & 78.78           & \textbf{97.15} \\ \hline
3                      & 65.86 & 75.42 & 65.71 & 74.82          & 65.52  & 74.02 & 72.02 & 94.51           & 75.36           & 71.34 & 90.37          & 95.27           & 92.37           & \textbf{97.95} \\ \hline
4                      & 73.71 & 58.12 & 64.10 & 49.32          & 51.31  & 71.49 & 55.87 & 79.34           & 64.58           & 74.31 & 86.90          & \textbf{99.22}  & 97.34           & 95.62          \\ \hline
5                      & 90.32 & 95.17 & 94.57 & 95.35          & 87.70  & 90.11 & 93.09 & 93.08           & 88.81           & 97.95 & 94.25          & 95.17           & 93.87           & \textbf{96.09} \\ \hline
6                      & 92.09 & 96.12 & 98.39 & 95.59          & 95.10  & 94.06 & 94.67 & 96.34           & 87.00           & 96.44 & 97.07          & \textbf{98.73}  & 98.27           & 96.80          \\ \hline
7                      & 96.00 & 11.53 & 50.00 & 36.92          & 56.92  & 84.61 & 80.00 & 76.00           & 90.00           & 54.02 & 91.26          & 96.00           & \textbf{98.67}  & 70.67          \\ \hline
8                      & 98.14 & 93.87 & 99.80 & 99.75          & 96.64  & 98.37 & 98.37 & 99.76           & 89.72           & 88.99 & 97.79          & \textbf{99.84}  & 99.77           & 98.68          \\ \hline
9                      & 11.11 & 0.00  & 0.00  & 5.00           & 28.89  & 33.33 & 88.89 & \textbf{100.00} & \textbf{100.00} & 58.89 & 75.91          & 96.30           & 98.15           & 70.37          \\ \hline
10                     & 73.80 & 80.89 & 59.30 & 69.11          & 75.12  & 86.05 & 74.49 & 79.51           & 77.19           & 73.10 & 87.34          & 87.01           & 78.86           & \textbf{97.45} \\ \hline
11                     & 55.41 & 83.02 & 84.44 & 89.91          & 83.49  & 82.98 & 69.58 & 85.42           & 77.58           & 70.78 & 90.24          & 89.08           & 81.75           & \textbf{98.40} \\ \hline
12                     & 76.92 & 86.32 & 77.04 & 86.78          & 67.55  & 73.40 & 65.29 & 84.24           & 72.00           & 57.51 & \textbf{95.76} & 93.51           & 90.64           & 93.57          \\ \hline
13                     & 91.30 & 79.25 & 99.09 & \textbf{99.51} & 96.86  & 87.02 & 98.37 & 98.91           & 87.80           & 99.25 & 97.49          & 98.56           & 98.56           & 95.32          \\ \hline
14                     & 93.32 & 88.87 & 92.50 & 96.45          & 96.51  & 94.38 & 91.39 & 98.06           & 93.48           & 95.07 & 96.03          & 95.73           & 93.24           & \textbf{98.51} \\ \hline
15                     & 67.72 & 60.00 & 68.68 & 61.79          & 39.08  & 75.57 & 48.99 & 87.31           & 72.36           & 91.26 & 90.48          & \textbf{97.31}  & 97.02           & 96.73          \\ \hline
16                     & 90.36 & 53.68 & 85.26 & 84.16          & 89.40  & 79.76 & 87.95 & 96.38           & 97.22           & 91.27 & 98.82          & 98.02           & \textbf{98.81}  & 79.76          \\ \hline
\textbf{AA(\%)}        & 76.89 & 66.72 & 72.65 & 73.10          & 71.66  & 84.44 & 76.89 & 89.31           & 81.18           & 77.43 & 92.04          & \textbf{95.09}  & 93.51           & 91.78          \\ \hline
\textbf{OA(\%)}        & 76.88 & 81.79 & 80.05 & 85.19          & 79.66  & 80.72 & 76.88 & 88.55           & 79.78           & 77.18 & 92.35          & 91.80           & 88.03           & \textbf{97.08} \\ \hline
\end{tabular}
\end{table*}

The results in the SA dataset are shown in Tab. \ref{tab:SA}. Similar to the IP dataset results, AAE performed best in five classes, VAE and ContrastNet both performed best in three classes. Though all methods got good performances in this dataset, ContrastNet still got the highest OA and got the highest AA. Unlike the IP dataset, the number of samples in the SA dataset is relatively balanced, so ContrastNet got a higher AA than AAE and VAE. The same conclusion can be drawn from the results in the PU dataset, which are shown in Tab.\ref{tab:PU}. ContrastNet still outperformed any other methods in terms of OA and AA. 

\begin{table*}[htpb]
\caption{Comparison in the Salinas dataset. 5\% labeled samples are used for training, and the rest are used for testing. AA menas average accuracy and OA means overall accuracy. The best values in each row are in bold. And the methods in bold are structures proposed in this paper. Some of the data in the table is quoted from \cite{3DCAE}.}
\label{tab:SA}
\begin{tabular}{c|cccccc|cccccccc}
\hline
Class  & \multicolumn{6}{c|}{Supervised Feature Extraction}                          & \multicolumn{8}{c}{Unsupervised Feature Extraction}                                                                     \\ \hline
       & LDA            & LFDA  & SGDA            & SLGDA           & 1D-CNN & S-CNN & PCA   & TPCA            & SSAE            & EPLS  & 3DCAE           & \textbf{AAE}             & \textbf{VAE}             & \textbf{ContrastNet}    \\ \hline
1      & 99.16          & 99.20 & 99.65           & 98.14           & 97.98  & 99.55 & 97.48 & 99.88           & \textbf{100.00} & 99.99 & \textbf{100.00} & 99.98           & 99.74           & 99.93          \\ \hline
2      & 99.94          & 99.19 & 99.33           & 99.44           & 99.25  & 99.43 & 99.52 & 99.49           & 99.52           & 99.92 & 99.29           & \textbf{100.00} & 99.79           & 99.80          \\ \hline
3      & 99.79          & 99.75 & 99.30           & 99.29           & 94.43  & 98.81 & 99.41 & 99.04           & 94.24           & 98.75 & 97.13           & \textbf{100.00} & \textbf{100.00} & 99.95          \\ \hline
4      & 99.77          & 99.71 & 99.21           & 99.57           & 99.42  & 97.45 & 99.77 & \textbf{99.84}  & 99.17           & 98.52 & 97.91           & 99.09           & 99.57           & 98.01          \\ \hline
5      & 98.98          & 98.47 & 99.07           & 98.06           & 96.60  & 97.96 & 98.70 & 98.96           & 98.82           & 98.33 & 98.26           & 99.42           & \textbf{99.71}  & 99.48          \\ \hline
6      & 99.89          & 99.09 & 99.57           & 99.32           & 99.51  & 99.83 & 99.65 & 99.80           & \textbf{100.00} & 99.92 & 99.98           & 99.97           & 99.86           & 99.94          \\ \hline
7      & \textbf{99.97} & 99.66 & 99.27           & 99.33           & 99.27  & 99.59 & 99.94 & 99.84           & 99.94           & 97.69 & 99.64           & 99.93           & 99.96           & 99.79          \\ \hline
8      & 81.84          & 86.89 & 89.78           & 89.48           & 86.79  & 94.40 & 83.90 & 84.11           & 80.73           & 78.86 & 91.58           & 91.21           & 86.91           & \textbf{99.53} \\ \hline
9      & 99.90          & 97.34 & \textbf{100.00} & 99.65           & 99.08  & 98.85 & 99.97 & 99.60           & 99.47           & 99.54 & 99.28           & 99.69           & 99.99           & 99.71          \\ \hline
10     & 96.31          & 95.85 & 97.99           & 97.94           & 93.71  & 97.35 & 96.89 & 95.76           & 92.12           & 95.98 & 96.65           & 98.46           & 96.54           & \textbf{99.80} \\ \hline
11     & 99.61          & 97.94 & 99.53           & 99.06           & 94.55  & 97.71 & 96.84 & 96.14           & 96.62           & 98.60 & 97.74           & 99.57           & \textbf{100.00} & 99.80          \\ \hline
12     & 99.67          & 99.64 & \textbf{100.00} & \textbf{100.00} & 99.59  & 98.73 & 99.95 & 99.07           & 97.75           & 99.44 & 98.84           & \textbf{100.00} & 99.44           & 99.98          \\ \hline
13     & 99.20          & 97.38 & 98.47           & 97.82           & 97.50  & 96.72 & 99.54 & \textbf{100.00} & 95.81           & 98.85 & 99.26           & 99.89           & 97.24           & 98.20          \\ \hline
14     & 96.56          & 93.18 & 96.07           & 90.47           & 94.08  & 95.22 & 97.24 & 95.74           & 96.65           & 98.56 & 97.49           & \textbf{99.08}  & 96.49           & 98.62          \\ \hline
15     & 73.60          & 63.04 & 65.40           & 69.51           & 66.52  & 95.61 & 76.68 & 79.54           & 79.73           & 83.13 & 87.85           & 93.69           & 87.90           & \textbf{99.53} \\ \hline
16     & 98.48          & 99.00 & 99.34           & 99.00           & 97.48  & 99.44 & 97.90 & 98.40           & 99.12           & 99.50 & 98.34           & \textbf{99.73}  & 99.61           & 99.57          \\ \hline
AA(\%) & 96.42          & 95.33 & 96.37           & 96.01           & 94.73  & 97.39 & 96.46 & 93.24           & 95.61           & 96.55 & 97.45           & 98.73           & 97.67           & \textbf{99.48} \\ \hline
OA(\%) & 92.18          & 91.22 & 91.82           & 93.31           & 91.30  & 97.92 & 92.87 & 96.57           & 92.11           & 92.35 & 95.81           & 97.10           & 95.23           & \textbf{99.60} \\ \hline
\end{tabular}
\end{table*}

\begin{table*}[htpb]
\caption{Comparison in the University of Pavia dataset. 10\% labeled samples are used for training, and the rest are used for testing. AA menas average accuracy and OA means overall accuracy. The best values in each row are in bold. And the methods in bold are structures proposed in this paper. Some of the data in the table is quoted from \cite{3DCAE}.}
\label{tab:PU}
\begin{tabular}{c|cccccc|cccccccc}
\hline
Class  & \multicolumn{6}{c|}{Supervised Feature Extraction}       & \multicolumn{8}{c}{Unsupervised Feature Extraction}                                                                             \\ \hline
       & LDA             & LFDA  & SGDA  & SLGDA & 1D-CNN & S-CNN & PCA             & TPCA            & SSAE            & EPLS           & 3DCAE  & \textbf{AAE}             & \textbf{VAE}             & \textbf{ContrastNet}    \\ \hline
1      & 78.41           & 93.45 & 91.37 & 94.66 & 90.93  & 95.40 & 90.89           & 96.17           & 95.72           & 95.95          & 95.21 & 95.39           & 81.31          & \textbf{99.49} \\ \hline
2      & 83.69           & 97.36 & 97.23 & 97.83 & 96.94  & 97.31 & 93.27           & 97.95           & 94.13           & 95.91          & 96.06 & 98.96           & 97.65          & \textbf{99.98} \\ \hline
3      & 73.02           & 71.41 & 66.08 & 77.27 & 69.43  & 81.21 & 82.60           & 86.50           & 87.47           & 94.33          & 91.32 & 97.49           & 91.00          & \textbf{99.06} \\ \hline
4      & 93.68           & 91.00 & 91.19 & 93.18 & 90.32  & 95.83 & 92.41           & 94.84           & 96.91           & \textbf{99.28} & 98.28 & 96.94           & 94.79          & 97.75          \\ \hline
5      & \textbf{100.00} & 97.99 & 99.33 & 98.51 & 99.44  & 99.91 & 98.98           & \textbf{100.00} & 99.76           & 99.92          & 95.55 & 99.94           & 99.94          & 99.81          \\ \hline
6      & 88.51           & 87.55 & 80.31 & 90.08 & 73.69  & 95.29 & 92.00           & 94.76           & 95.76           & 93.57          & 95.30 & 99.29           & \textbf{99.91} & 99.90          \\ \hline
7      & 85.75           & 80.23 & 75.26 & 85.34 & 83.42  & 87.05 & 85.83           & 91.89           & 91.18           & 98.17          & 95.14 & \textbf{100.00} & 99.42          & 99.83          \\ \hline
8      & 74.49           & 86.98 & 84.11 & 90.49 & 83.65  & 87.35 & 82.96           & 89.04           & 82.47           & 91.23          & 91.38 & 97.90           & 94.77          & \textbf{98.79} \\ \hline
9      & 99.11           & 99.26 & 95.14 & 99.37 & 98.23  & 95.66 & \textbf{100.00} & 98.94           & \textbf{100.00} & 99.78          & 99.96 & 96.32           & 88.62          & 94.84          \\ \hline
AA(\%) & 86.29           & 89.47 & 86.67 & 91.86 & 87.34  & 94.75 & 90.99           & 95.64           & 93.71           & 96.33          & 95.36 & 98.03           & 94.16          & \textbf{98.83} \\ \hline
OA(\%) & 83.75           & 92.77 & 90.58 & 94.15 & 89.99  & 92.78 & 91.37           & 94.45           & 93.51           & 95.13          & 95.39 & 98.14           & 94.53          & \textbf{99.46} \\ \hline
\end{tabular}
\end{table*}

Generally speaking, the proposed AAE, VAE, and ContrastNet in this paper are good models that outperform many supervised and unsupervised methods. Among the three methods, Autoencoder-based AAE and VAE are models that make use of reconstruction information and concentrate on the single sample itself. ContrastNet is a model that learns the similarity between two different distributions of a single sample, and also learn the dissimilarity between positive samples and negative samples. According to the experiment results, ContrastNet performs better than AAE and VAE in OA and AA when the number of samples is balanced. When the number of samples is imbalanced, ContrastNet tends to get higher OA and lower AA than AAE and VAE.

\begin{figure*}[htpb]
\centering
\subfloat[AAE]{\includegraphics[width=2.5in]{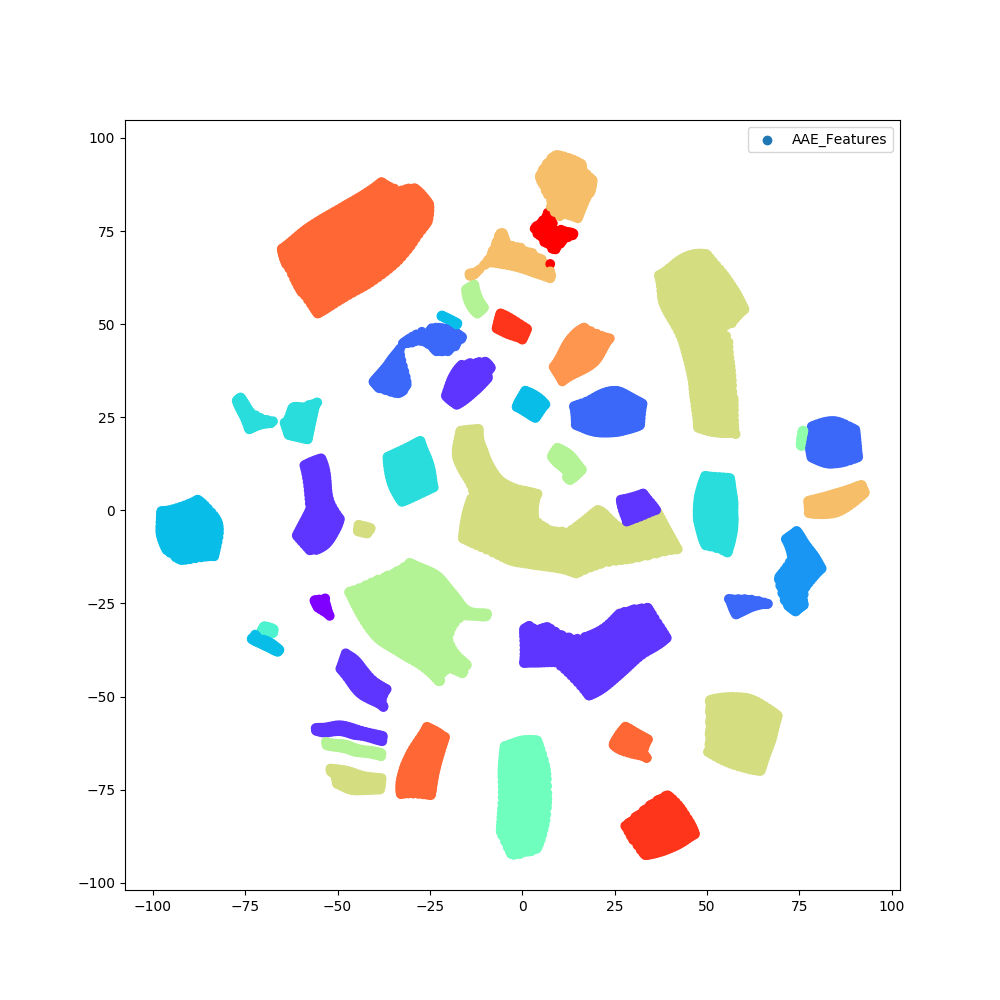}
\label{fig:tsne_IP_AAE}}
\subfloat[VAE]{\includegraphics[width=2.5in]{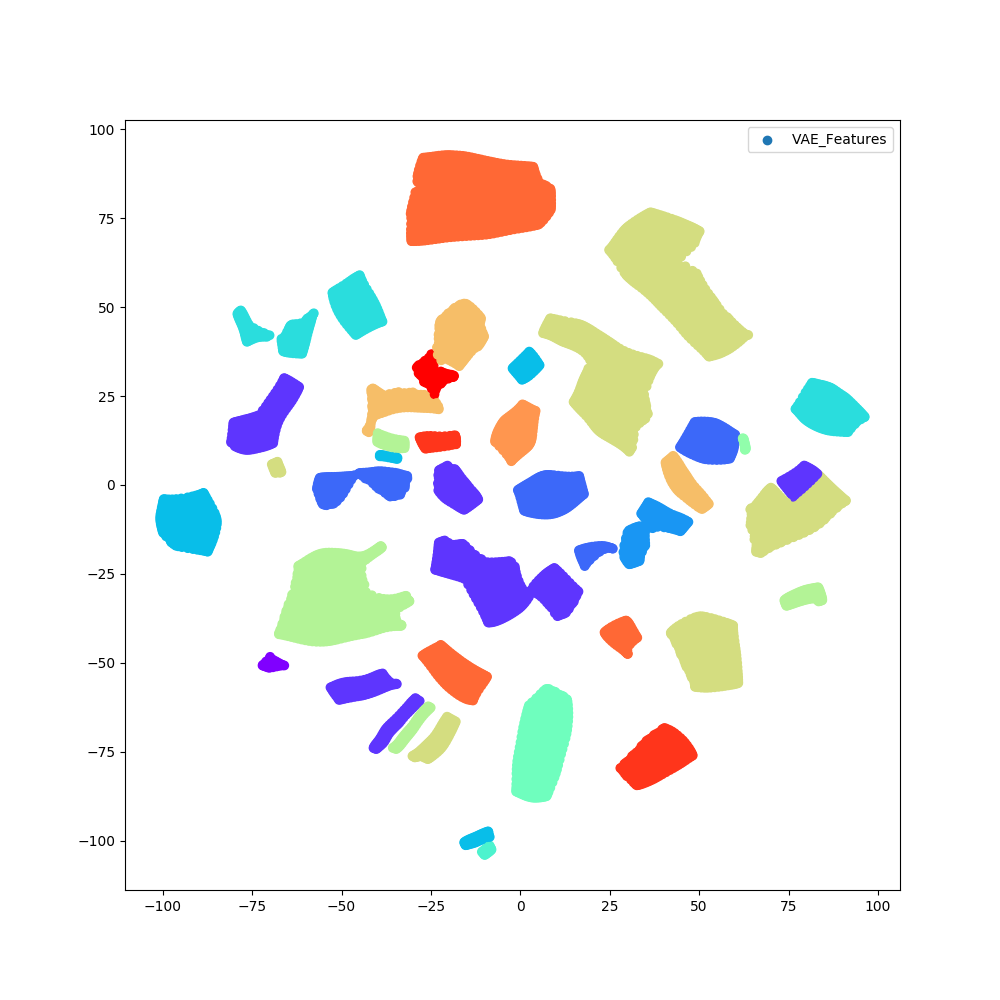}
\label{fig:tsne_IP_VAE}}
\subfloat[ContrastNet]{\includegraphics[width=2.5in]{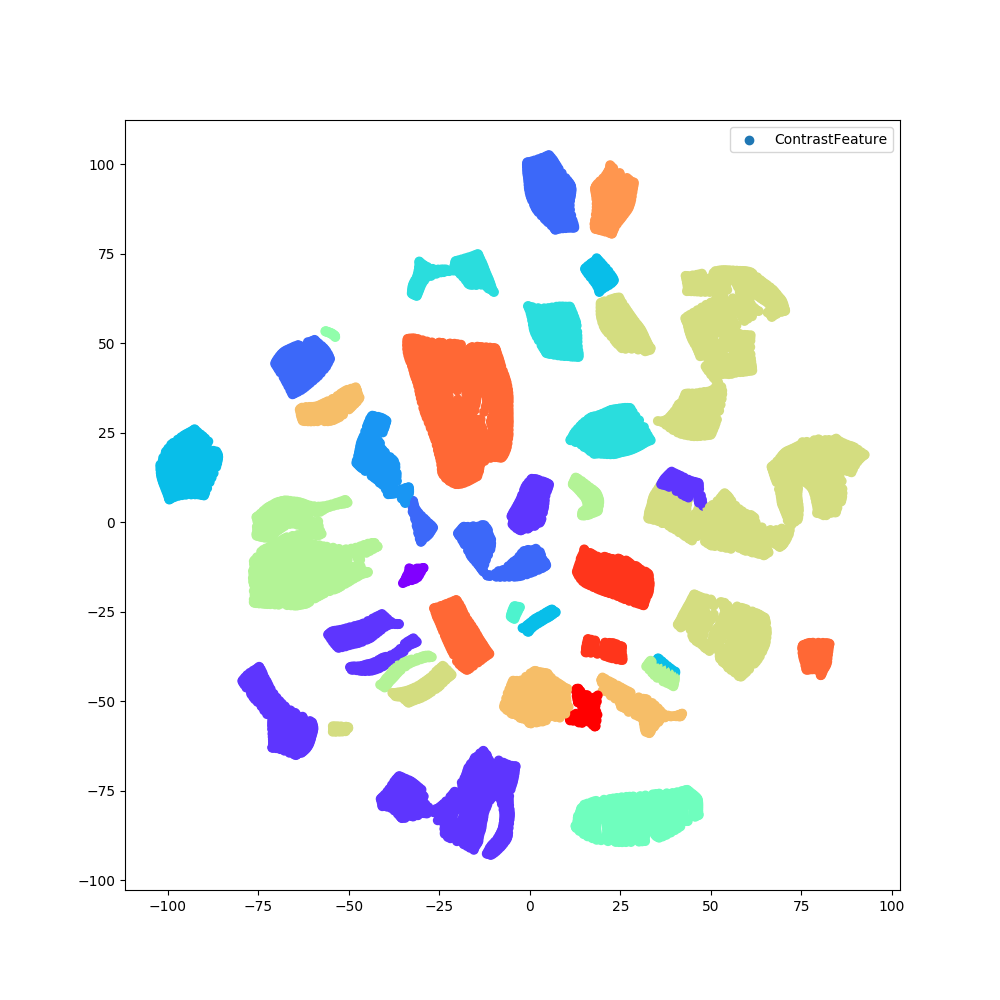}%
\label{fig:tsne_IP_Contrast}}
\caption{Visualization of extracted features in the Indian Pines dataset. Each class is corresponding to a kind of color.}
\label{fig:tsne_IP}
\end{figure*}

\begin{figure*}[htpb]
\centering
\subfloat[AAE]{\includegraphics[width=2.5in]{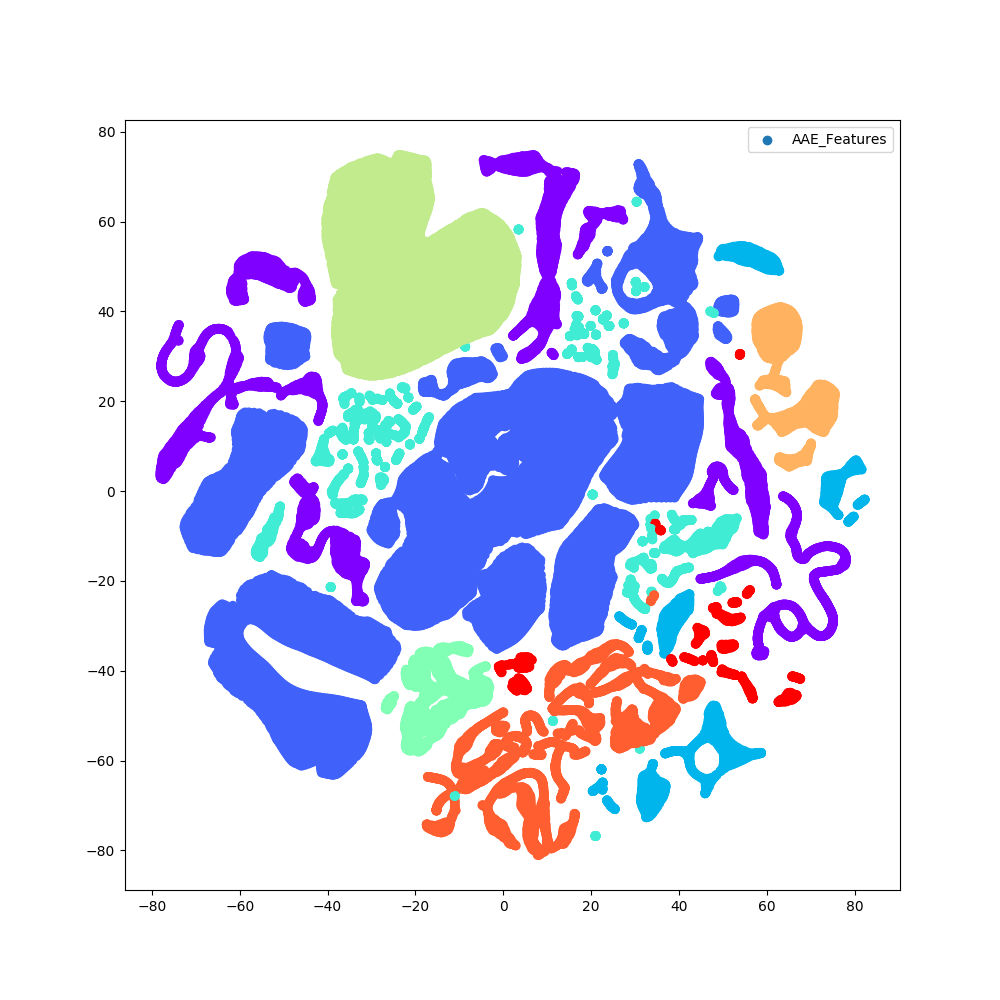}
\label{fig:tsne_PU_AAE}}
\subfloat[VAE]{\includegraphics[width=2.5in]{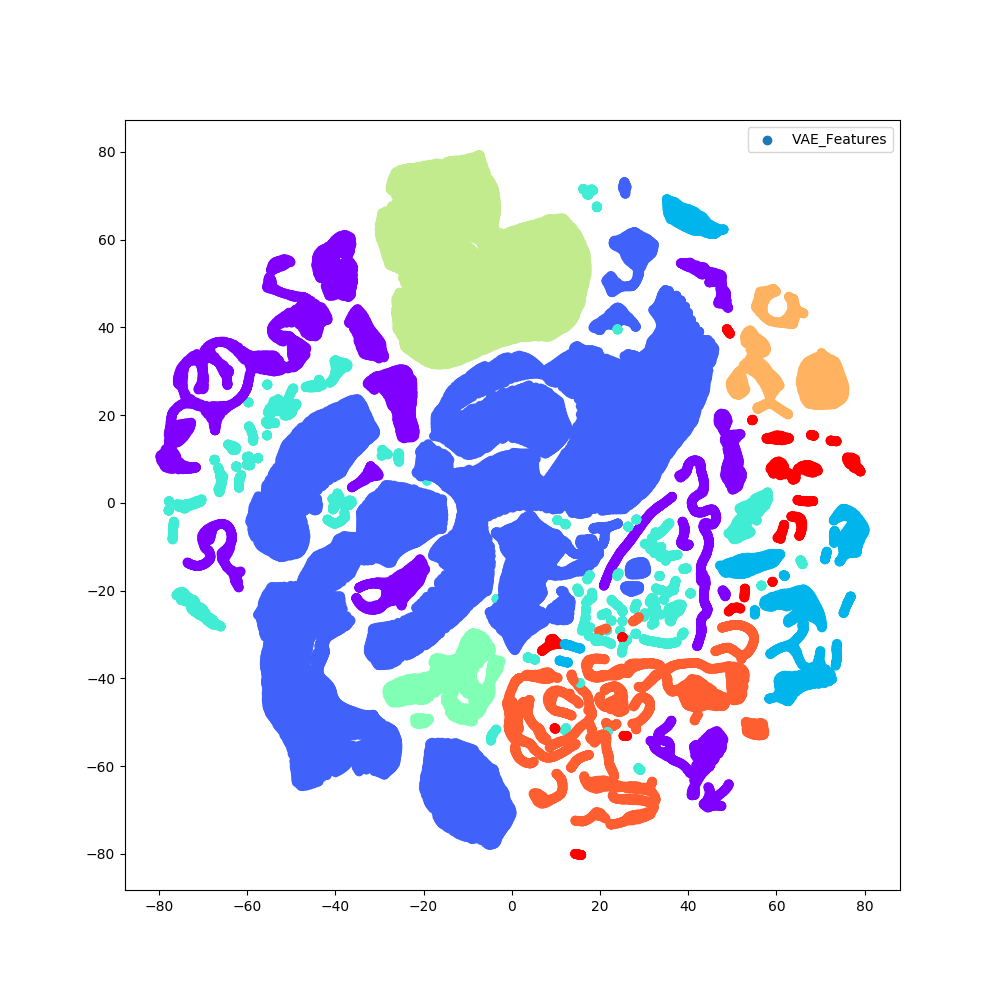}
\label{fig:tsne_PU_VAE}}
\subfloat[ContrastNet]{\includegraphics[width=2.5in]{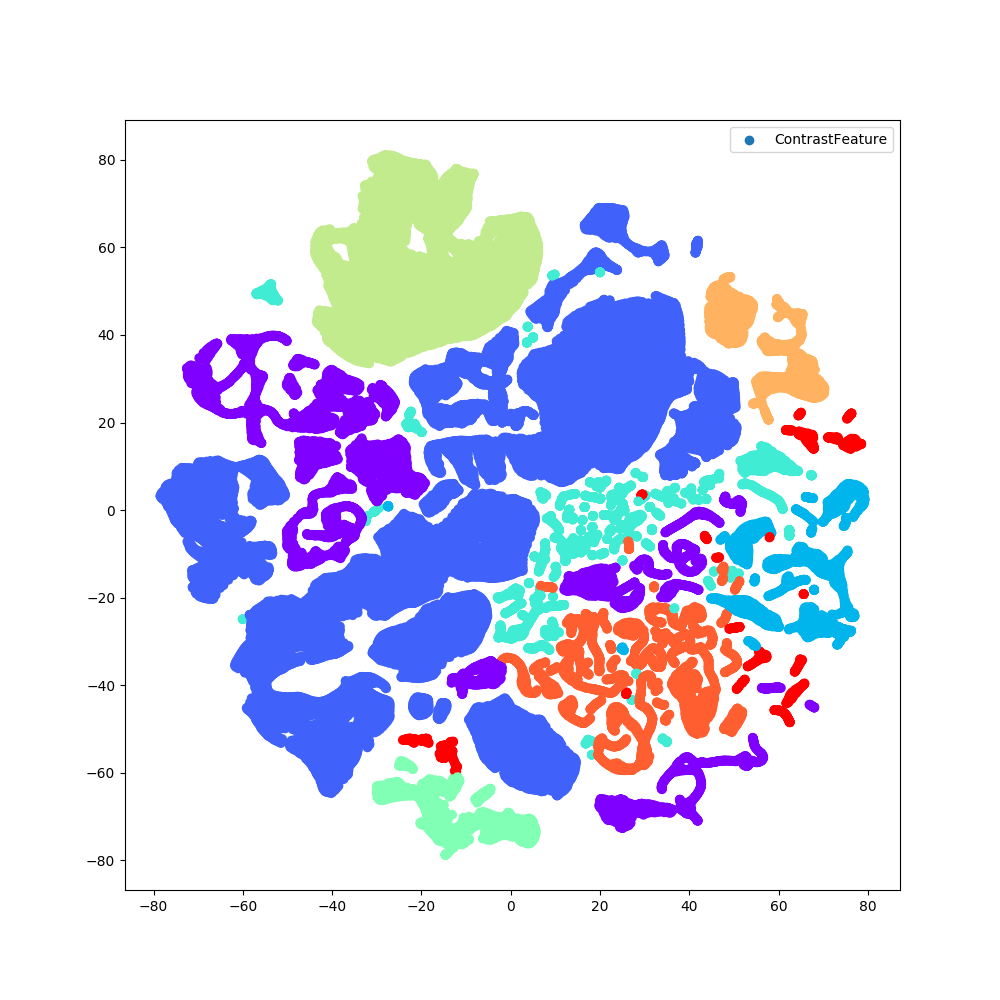}%
\label{fig:tsne_PU_Contrast}}
\caption{Visualization of extracted features in the University of Pavia dataset. Each class is corresponding to a kind of color.}
\label{fig:tsne_PU}
\end{figure*}

\begin{figure*}[htpb]
\centering
\subfloat[AAE]{\includegraphics[width=2.5in]{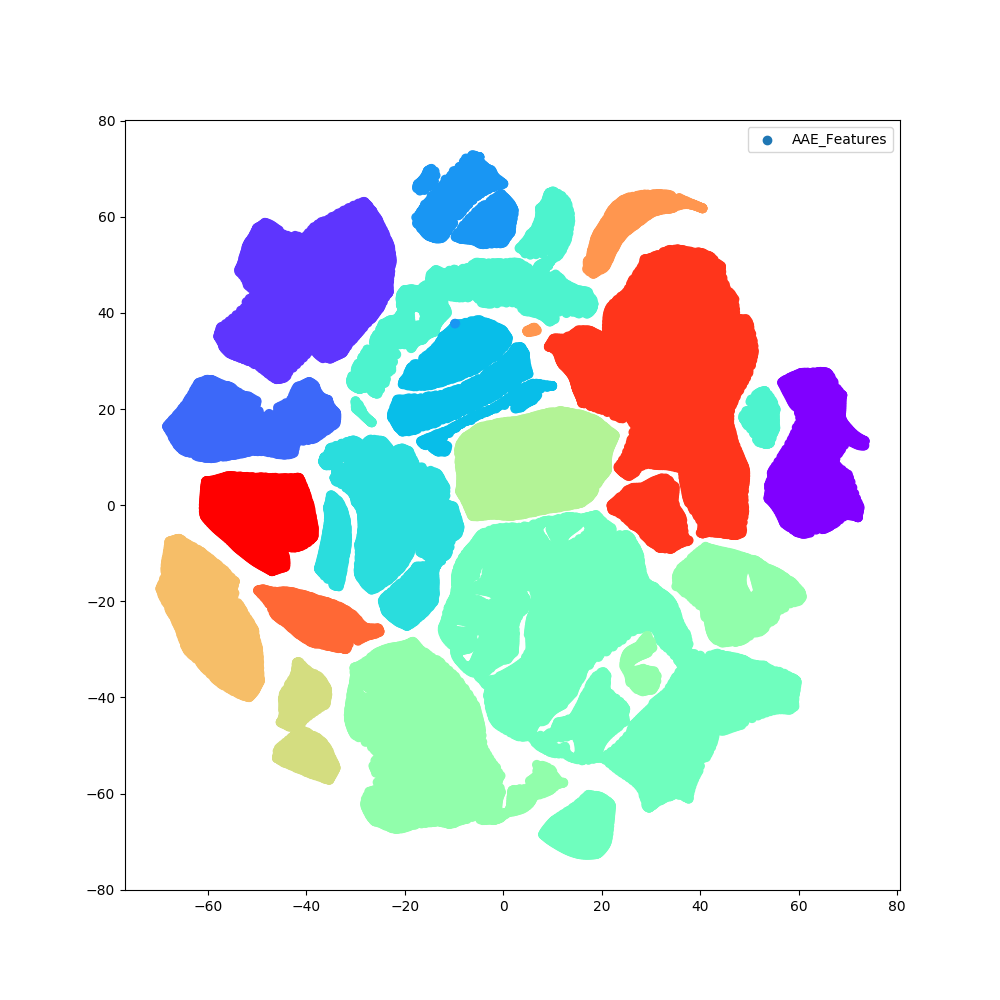}
\label{fig:tsne_SA_AAE}}
\subfloat[VAE]{\includegraphics[width=2.5in]{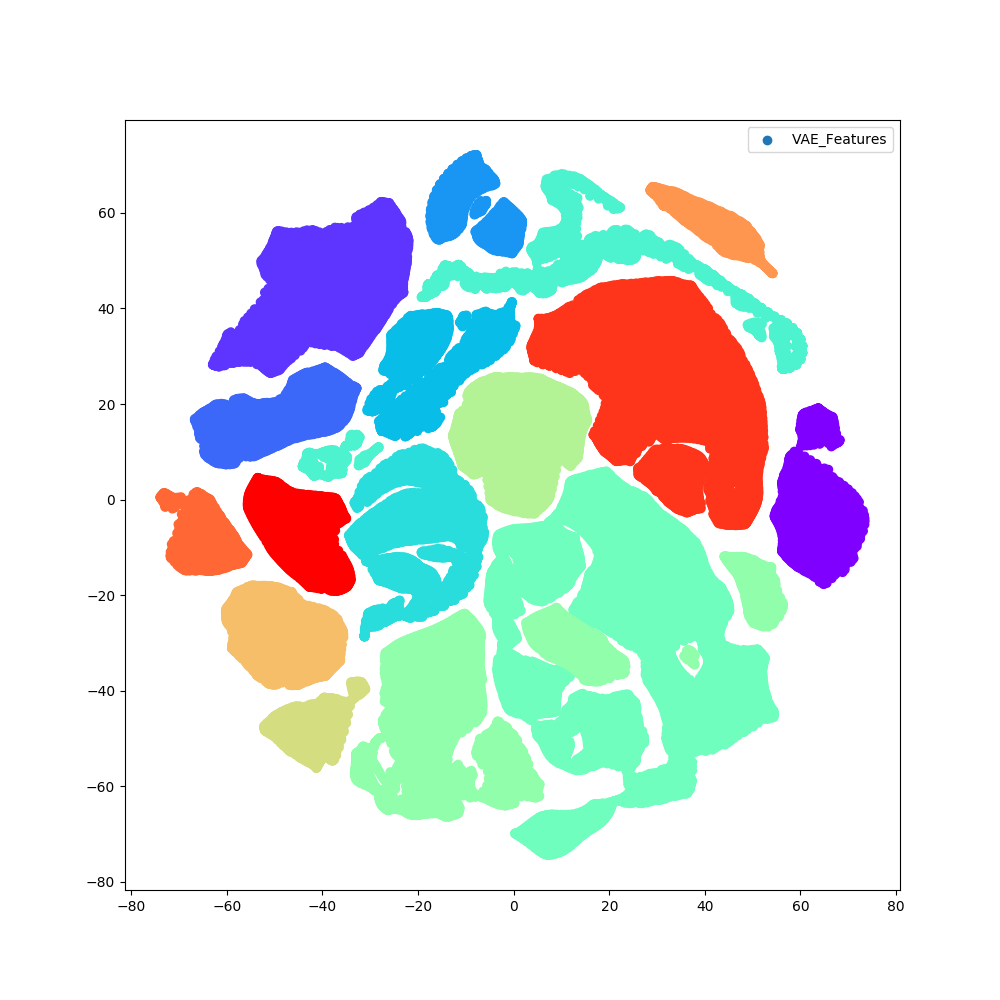}
\label{fig:tsne_SA_VAE}}
\subfloat[ContrastNet]{\includegraphics[width=2.5in]{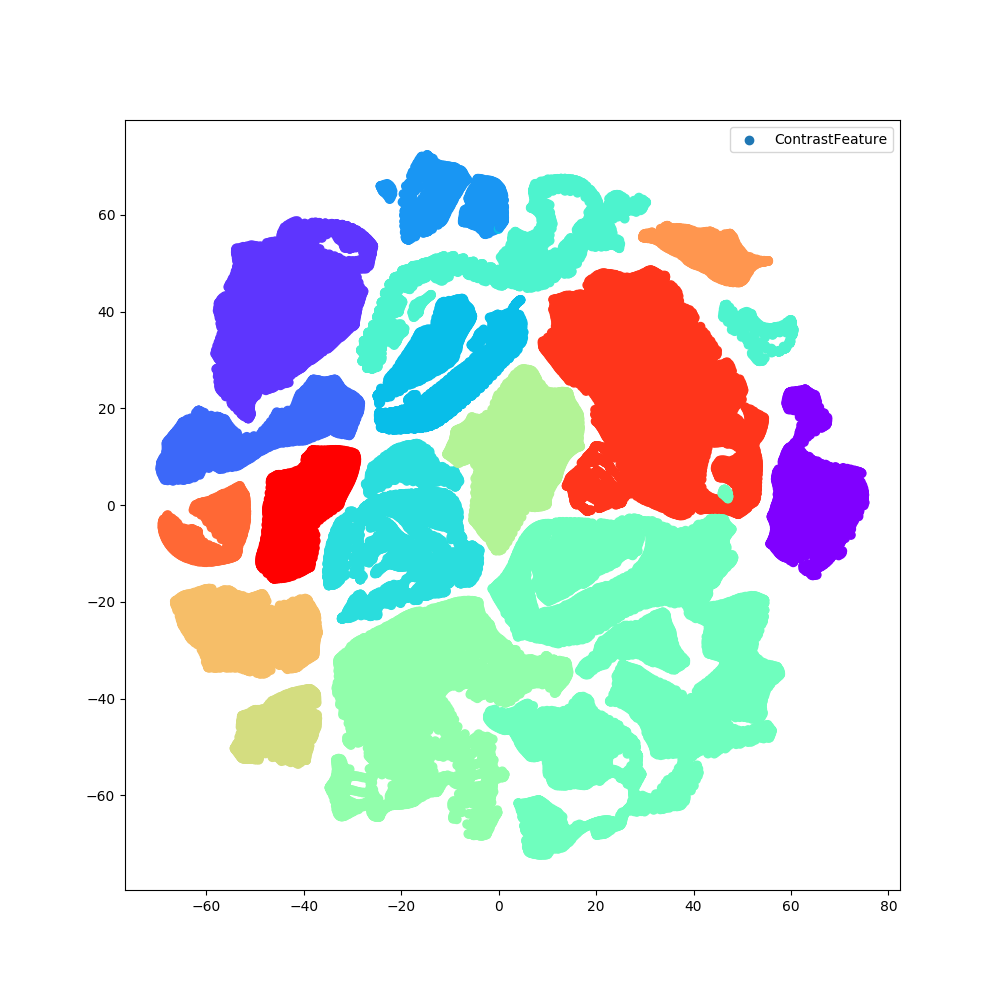}%
\label{fig:tsne_SA_Contrast}}
\caption{Visualization of extracted features in the Salinas dataset. Each class is corresponding to a kind of color.}
\label{fig:tsne_SA}
\end{figure*}

To intuitively show the clustering effect of ContrastNet, we use t-SNE\cite{TSNE} method to visualize the features extracted by AAE, VAE, and ContrastNet. The visualizations are shown in Fig. \ref{fig:tsne_IP}, \ref{fig:tsne_PU}, \ref{fig:tsne_SA}. It is not hard to find that the features extracted by the ContrastNet are easier to classify. Because the inner-class distances are small, and the inter-class distances are big in the ContrastNet features.

The computational analysis is shown in Tab. \ref{tab:time}. Because the training of ContrastNet demands extracted AAE and VAE features, which means the demand for AAE and VAE, the training time of ContrastNet is considerably longer than other algorithms. However, When extracting features, the time consumption of ContrastNet is fairly fast than others. Because for AAE and VAE, only encoders are used in the feature extraction phase. Moreover, the structure of   AAE and VAE is relatively simple and efficient, so the ContrastNet performs fast in the feature extraction phase.

\begin{table*}[htpb]
\caption{Computational analysis of different feature extration algorithms. The best value in each row is in bold. And the methods in bold are structures proposed in this paper. Some of the data in the table is quoted from \cite{3DCAE}.}
\label{tab:time}
\centering
\begin{tabular}{cccccccc}
\hline
\multicolumn{8}{c}{Indian Pines dataset}                                                                                     \\ \hline
\multicolumn{1}{c|}{}                       & 1D-CNN        & S-CNN & TPCA & SSAE          & EPLS   & 3DCAE & \textbf{ContrastNet}    \\ \hline
\multicolumn{1}{c|}{Training(s)}            & \textbf{20.6} & 119   & 44   & 240.1         & 111.9  & 1156  & 2073           \\ \hline
\multicolumn{1}{c|}{Feature Extraction (s)} & 17.4          & 3.2   & 150  & \textbf{2.76} & 47.3   & 5.22  & 12.77          \\ \hline
\multicolumn{8}{c}{Salinas dataset}                                                                                   \\ \hline
\multicolumn{1}{c|}{}                       & 1D-CNN        & S-CNN & TPCA & SSAE          & EPLS   & 3DCAE & \textbf{ContrastNet}    \\ \hline
\multicolumn{1}{c|}{Training(s)}            & \textbf{43.1} & 134   & 241  & 513           & 103.5  & 1159  & 4784           \\ \hline
\multicolumn{1}{c|}{Feature Extraction (s)} & 83.8          & 62    & 816  & \textbf{16.3} & 192.2  & 26.4  & 36.65          \\ \hline
\multicolumn{8}{c}{University of Pavia dataset}                                                                                 \\ \hline
\multicolumn{1}{c|}{}                       & 1D-CNN        & S-CNN & TPCA & SSAE          & EPLS   & 3DCAE & \textbf{ContrastNet}    \\ \hline
\multicolumn{1}{c|}{Training(s)}            & \textbf{32.1} & 332   & 44   & 491           & 127.13 & 1168  & 3778           \\ \hline
\multicolumn{1}{c|}{Feature Extraction (s)} & 150.07        & 106   & 150  & 31.9          & 141.5  & 32.04 & \textbf{27.82} \\ \hline
\end{tabular}
\end{table*}
\section{Conclusion}
In this paper, we proposed an unsupervised feature learning method based on autoencoder and contrastive learning. This method combines unsupervised representative methods and unsupervised discriminative methods, learning to extract better hyperspectral classification features than other baseline methods. In the proposed method, we designed two efficient autoencoder structure: VAE and AAE, and design a ContrastNet for contrastive learning, which reduces the computing resources demand of contrastive learning. Our experiments show that the proposed method can extract more representative features and keep a high feature extraction speed in the testing phase. Our work shows that unsupervised learning still has great potential in the remote sensing field, and we hope others can get more exciting ideas through our exploration.

% or


\appendices
\section{}

\section*{Acknowledgment}

The authors would like to thank...

\ifCLASSOPTIONcaptionsoff
  \newpage
\fi




\bibliographystyle{IEEEtran}
\bibliography{article}


%\end{IEEEbiography}

\begin{IEEEbiographynophoto}{Zeyu Cao}
received the B.S. degree in automation from Zhejiang University, Hangzhou, China, where he is currently pursuing the Ph.D. degree in control theory and control engineering. His research interests include object detection and machine learning.
\end{IEEEbiographynophoto}

%\newpage

\begin{IEEEbiographynophoto}{Xiaorun Li}
received the B.S. degree from the National University of Defense Technology, Changsha, China, in 1992, and the M.S. and Ph.D. degrees from Zhejiang University, Hangzhou, China, in 1995 and 2008, respectively.
Since 1995, he has been with Zhejiang University, where he is currently a Professor with the College of Electrical Engineering. His research interests include hyperspectral image processing, signal and image processing, and pattern recognition.
\end{IEEEbiographynophoto}

\begin{IEEEbiographynophoto}{Liaoying Zhao}
received the B.S. and M.S. degrees from Hangzhou Dianzi University, Hangzhou, China, in 1992 and 1995, respectively, and the Ph.D. degree from Zhejiang University, Hangzhou, in 2004.
Since 1995, she has been with Hangzhou Dianzi University, where she is currently a Professor with the College of Computer Science. Her research interests include hyperspectral image processing, signal and image processing, pattern recognition, and machine learning.
\end{IEEEbiographynophoto}


%\vfill


\end{document}